\documentclass[runningheads]{llncs}

\usepackage{eccv}

\usepackage{eccvabbrv}

\usepackage{graphicx}
\usepackage{booktabs}
\usepackage{comment}

\usepackage[accsupp]{axessibility}  

\usepackage[pagebackref,breaklinks,colorlinks,citecolor=eccvblue]{hyperref}

\usepackage{orcidlink}

\begin{document}

\title{Progressive Monitoring of Generative Model Training Evolution}

\titlerunning{Progressive Monitoring of DGM Training}

\author{Vidya Prasad \and Anna Vilanova \and
Nicola Pezzotti}

\authorrunning{Prasad et al.}

\institute{Eindhoven University of Technology, Netherlands\\
\email{v.prasad@tue.nl}
}

\maketitle

\begin{abstract}

While deep generative models (DGMs) have gained popularity, their susceptibility to biases and other inefficiencies that lead to undesirable outcomes remains an issue. 
With their growing complexity, there is a critical need for early detection of issues to achieve desired results and optimize resources. 
Hence, we introduce a progressive analysis framework to monitor the training process of DGMs. 
Our method utilizes dimensionality reduction techniques to facilitate the inspection of latent representations, the generated and real distributions, and their evolution across training iterations. 
This monitoring allows us to pause and fix the training method if the representations or distributions progress undesirably. 
This approach allows analysis of a model’s training dynamics and the timely identification of biases and failures, minimizing computational loads. 
We demonstrate how our method supports identifying and mitigating biases early in training a Generative Adversarial Network (GAN) and improving the quality of the generated data distribution.

\keywords{Generative Models \and Progressive Training Analysis \and Biases}
\end{abstract}

\section{Introduction}
\label{sec:intro}
In recent years, deep generative models (DGMs) have gained vast popularity and have revolutionized various domains, including computer vision~\cite{bond2021deep}. DGMs, such as variational autoencoders, generative adversarial networks (GANs), and diffusion models, learn to mimic the complex distributions of data, producing synthetic data indistinguishable from real datasets. 
Beyond the traditional image synthesis applications, these models have been applied to classification, anomaly detection, super-resolution, text-to-image conversion, and image-to-image translation~\cite{jin2020generative}, significantly improving the capabilities of vision models by generating high-fidelity, diverse datasets~\cite{bond2021deep}.

However, as these models evolve, their size and complexity increase~\cite{manduchi2024challenges}. 
This enhanced complexity introduces challenges in detecting bias and conducting error analysis~\cite{prasad2022transform}, which traditionally rely on manually analyzing aggregated performance metrics or selected samples~\cite{yu2020inclusive}. 
These methods are insufficient for a complete analysis within these complex models, limiting the ability to identify biases and errors effectively. 
This error analysis is particularly critical in sensitive applications such as handling human-centric datasets~\cite{luccioni2024stable}, where the accuracy and impartiality of the generated data are of utmost importance.
Furthermore, these complex models significantly increase training data size and the complexity of training processes~\cite{manduchi2024challenges, morales2024efficient}. 
This increase in complexity results in lengthy training processes, especially for industry-scale models~\cite{wu2022sustainable}. 

Prior works understand deep neural networks predominantly based on post-hoc analyses, exploring latent spaces~\cite{park2023content}, concepts~\cite{tcav}, causal units~\cite{bau2018visualizing} for specific outputs, features over model layers~\cite{yosinski2015understanding}, and generated data distributions~\cite{wang2018ganviz}. However, post-hoc analysis, rather than during training, can delay the detection of biases and errors, limit opportunities for model adjustment, and increase computational costs.
While systems to explore CNN training dynamics have been proposed~\cite{pezzotti2017deepeyes, conceptevo}, they typically focus on a per-iteration analysis and do not maintain a comprehensive view of the whole training process. 
This evolution context is crucial for understanding how biases and errors evolve and whether optimization proceeds undesirably. 
Despite advancements in literature, gaps remain in real-time bias detection and correction during training. 
Further, existing methodologies primarily concentrate on classification tasks or aggregate metrics, overlooking the analysis of high-dimensional data evolution during training.

We address these gaps in analyzing the evolution of high-dimensional model data, such as latent vectors, during training. 
Specifically, we need an interpretable method for tracking pattern changes and evolutions across training iterations (i.e., batches).
The ultimate goal is to support identifying and correcting undesired biased outcomes during the training phase rather than relying on post-hoc analysis.
Hence, we introduce a progressive analysis framework for real-time monitoring of the training evolution of DGMs. 
Model components, such as generated images and latent feature vectors, are systematically extracted at regular intervals. 
To visualize and analyze evolving data representations during model training, we propose adapting a recent evolutionary dimensionality reduction (DR) technique~\cite{prasad2024tree}. 
This method is used to project data from regular training intervals into a low-dimensional space and aligns it across iterations, facilitating the analysis of evolution.
This approach is similar to the common practice of monitoring loss metrics progressively, such as with TensorBoard, but instead facilitates a more holistic evaluation of the model's representations and data distribution across training iterations.
It enables early discovery of undesirable biases or inefficiencies, enabling timely adjustments to the training process.
Our framework's efficacy is demonstrated through empirical validation with a GAN to modify the hair color in human face datasets. 
It enables us to detect early signs of model problems and deviations from expected outcomes, save computational efforts, and facilitate transparent model development.

\section{Related work}
The interpretability of deep generative models (DGMs) is a challenging area due to their inherent complexity~\cite{manduchi2024challenges}. Traditional strategies predominantly rely on aggregate performance metrics, post-training analysis, and manual inspection of samples~\cite{zhou2021evaluation, yu2020inclusive, yang2023disdiff}. These approaches become increasingly inadequate as the scale and intricacy of modern DGMs grow.

A significant body of research has focused on post-hoc analysis to understand the inner workings of deep learning models~\cite{park2023content, tcav, wang2018ganviz}. 
Activation maximization techniques generate inputs that maximize specific neuron activations, offering insights into neuron roles across layers~\cite{yosinski2015understanding}. 
Inversion methods reconstruct input images from feature representations at various network layers, providing a holistic view of preserved information~\cite{yosinski2015understanding, pezzotti2017deepeyes}. 
In the context of DGMs, exploring causal units in GANs has facilitated controlled and targeted data generation beyond mere understanding~\cite{bau2018visualizing}. 
Moreover, the dynamics between synthetic and real datasets have been explored to enhance understanding~\cite{wang2018ganviz}. 
Disentanglement analysis within DGMs has further deepened insights into model biases~\cite{zhao2021codegan, yang2023disdiff, jeong2022interactively}. 
However, these analyses are typically conducted post-training, leading to inefficient resource utilization when issues are identified.

Interactive tools~\cite{pezzotti2017deepeyes, liu2017analyzing} have been developed to monitor the training process of models. DeepEyes~\cite{pezzotti2017deepeyes} employs progressive visual analytics techniques to provide real-time insights into the training process, identifying stable network layers and potential areas for improvement. Concept evolution methods have also been proposed to track the development of concepts within networks during training~\cite{conceptevo}. 
While these tools enable in-training analysis, they primarily focus on per-iteration analysis of high-dimensional data embeddings and internal feature maps~\cite{pezzotti2017deepeyes, conceptevo} for classification models or low-dimensional data such as loss metrics, gradients, and weights across iterations~\cite{liu2017analyzing} for DGMs. 
As a result, they often overlook the evolution patterns of high-dimensional data, especially in cases with large sample sets and numerous training iterations, where tracking and recalling earlier patterns becomes particularly challenging.

Despite these advancements, literature mainly addresses post-hoc or per-iteration analysis for classification models. There is a gap in the real-time exploration of high-dimensional data evolution during training, crucial for actively detecting and fixing biases in complex models like DGMs. Addressing this gap is essential for ensuring model quality and conserving computational resources.

\section{Background}
\label{sec:background}
\begin{figure}[t!]
    \centering
    \includegraphics[width=0.9\linewidth]{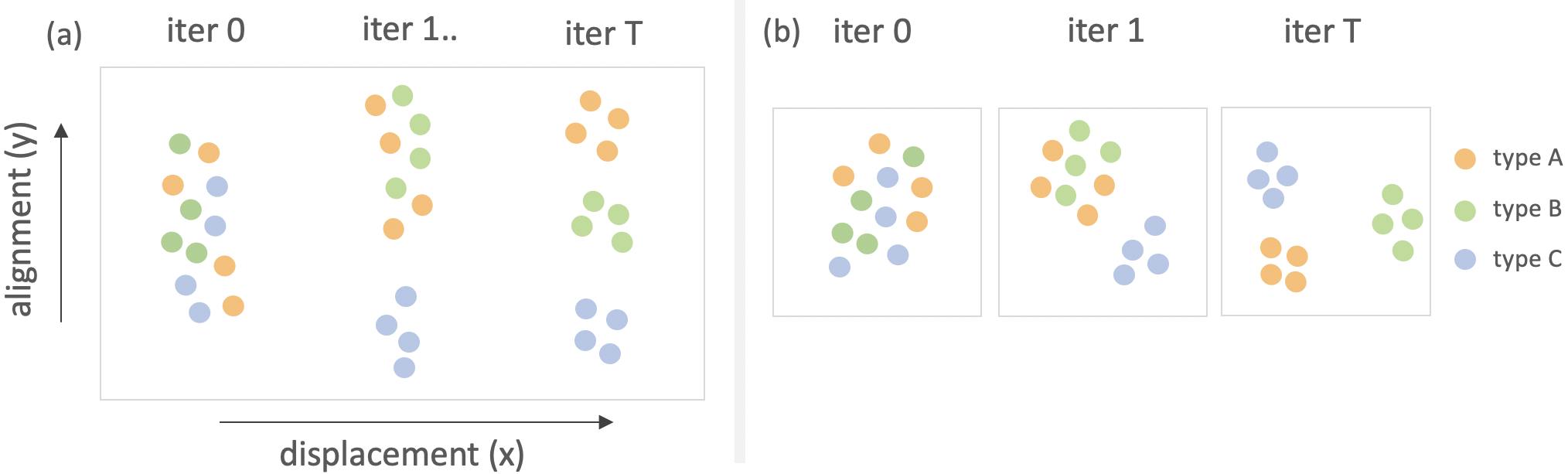}
\caption{(a) The evolutionary embedding method proposed by EvolvED~\cite{prasad2024tree} vs. (b) independent t-SNE~\cite{van2008visualizing} embeddings per iteration on example data. Each iteration is explicitly encoded and aligned with its prior step, enabling tracing evolutions in (a). }
    \label{fig:evoemb_desc}
\end{figure}

Non-linear DR methods, like t-distributed Stochastic Neighbor Embedding (t-SNE)~\cite{van2008visualizing} and Uniform Manifold Approximation and Projection (UMAP)~\cite{mcinnes2018umap}, are widely used to visualize high-dimensional data, like image distributions and latent feature vectors~\cite{wang2018ganviz}.
The primary objective of these methods is to maintain the neighborhood relationships of high-dimensional data points when mapped into a low-dimensional space.
However, analyzing the evolution of latent vectors across training iterations involves both high-dimensional data and iterative temporal aspects. 
Due to their stochasticity, these methods~\cite{van2008visualizing} fail to preserve the iterative structure essential for the analysis of pattern changes over iterations~\cite{prasad2024tree} (see Figure~\ref{fig:evoemb_desc}b).
To address this limitation, evolutionary embedding approaches have been recently introduced~\cite{prasad2024tree} (see Figure~\ref{fig:evoemb_desc}a) to maintain the iterative context, facilitating the visualization of data evolution. 

In addition to retaining the local neighborhood relationships per iteration, this method~\cite{prasad2024tree} enables detailed analysis of evolutionary processes by explicitly encoding and aligning each iteration with its preceding steps.
Each iteration is depicted as a distinct cluster of points horizontally grouped along a specific  displacement (vertical bands), differentiating each iteration spatially. 
Points from the same instance across different iterations are vertically aligned, facilitating analysis of their evolution over time. 
Points can also be color-encoded based on available metadata.
This explains how model components, such as latent vectors, evolve throughout training. In Figure~\ref{fig:evoemb_desc}a, initially mixed points at iteration 0 gradually form distinct clusters as training progresses, demonstrating the method's utility in exploring model evolution.
In contrast, studying evolutions with vanilla t-SNE (see Figure~\ref{fig:evoemb_desc}b) is more challenging. 
While the data in Figure~\ref{fig:evoemb_desc} is toy data, the pattern evolution tracing becomes very complex with more data points and many iterations (and, in turn, embeddings).

Initially developed to study the iterative generative process of trained diffusion models, this improved evolutionary embedding approach~\cite{prasad2024tree} has proven effective compared to vanilla t-SNE for analyzing high-dimensional evolutionary data. We propose using this approach to support the progressive analysis of deep learning (DL) model training processes, where "iteration" refers to a training batch, and we investigate the evolution of various model components.

\section{Progressive Framework}

\begin{figure}[t!]
    \centering
    \includegraphics[width=0.99\linewidth]{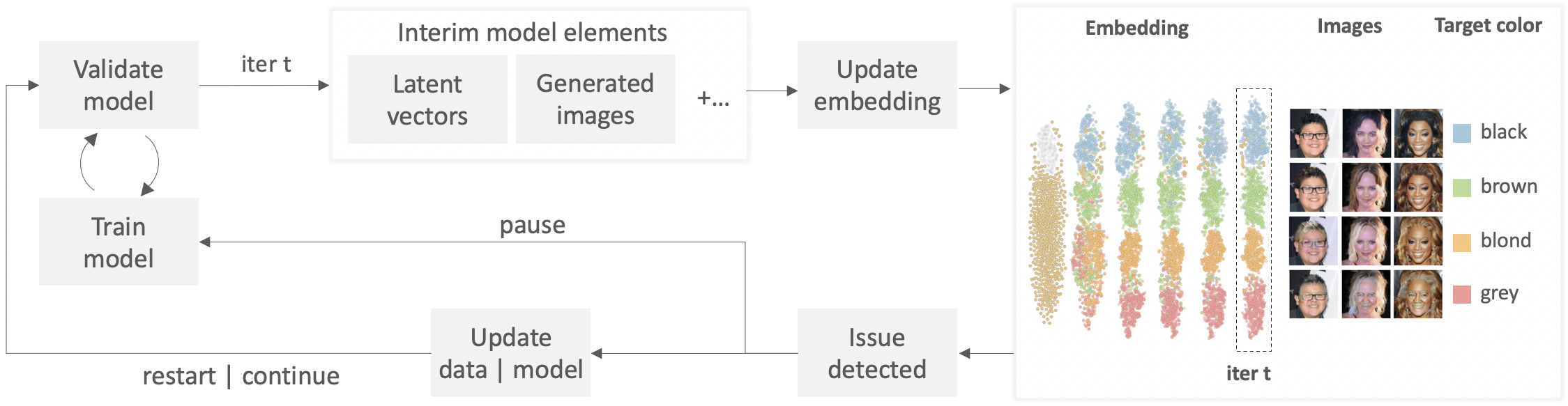}
    \caption{Overview of the progressive analysis framework. As training proceeds, model representations and generated datasets are extracted every $n$ iterations. These model elements are projected to a 2D space via DR methods that explicitly encode iterations and align each with the preceding steps. This embedding is explored with images to detect undesirable progression of the model. The training can be paused for timely corrections and resumed when issues are detected.}
    
    \label{fig:overview}
\end{figure}

We introduce a progressive analysis framework for real-time monitoring of DGM training. 
This approach focuses on preemptively identifying and mitigating biases and undesired results, enabling timely interventions rather than relying on post-hoc analysis. 
By integrating mid-training evaluations, our framework supports in-depth analysis of model representations and data distributions at specific intervals. 
This strategy reduces unnecessary training iterations, conserves computational resources, and enhances both training efficiency and model robustness.
\newline
\newline
\textbf{Workflow}: The workflow of the proposed framework is illustrated in Figure~\ref{fig:overview}. The training process of the DGM model begins with the initialization of model parameters and training data inputs as per the defined architecture. Unlike the traditional method of logging aggregate performance metrics (e.g., overall model loss) at every $n^{th}$ iteration (i.e., training batch), we extract and analyze key elements, including model representations (e.g., latent representations), and compare generated versus real data distributions. 
These elements logged at specific iterations are the model's training progression checkpoints. 
The alignment of these elements with desired outcomes is verified progressively for insight into the model's evolving capabilities and potential biases.
\newline
\newline
\textbf{DR Embedding}: We adapt a recently introduced evolutionary embedding method~\cite{prasad2024tree} (see Section~\ref{sec:background}) to progressively explore these extracted model elements. 
This method summarizes model elements extracted at every $n^{th}$ iteration and traces their evolution to the current iteration by aligning the embeddings across iterations. 
This provides an interpretable way to track pattern changes across the training process.
Figure~\ref{fig:overview} depicts the evolution of model elements, such as the discriminator's latent features.
The data is organized within a designated two-dimensional space at every $n^{th}$ iteration.  
An embedding for each model element is created, for example, an embedding of the discriminator's latent space or generated (and real) images. 

While latent space features can be projected directly into 2D using dimensionality reduction methods, effectively studying the semantics of high-dimensional images (e.g., real and fake images) requires going beyond pixel-level differences due to their large size and lack of semantic meaning at a pixel-level. 
Therefore, foundational image encoders, like CLIP~\cite{clip}, can be utilized. 
Images are projected into a latent space using this image encoder, and the evolutionary embedding method is then applied to obtain the projected 2D points for simplified analysis. 
Note that while we focus on CLIP as an example in our cases, any encoder suitable for the specific case can be utilized.
\newline
\newline
\textbf{Training Intervention}: 
Inspecting this embedding provides information about the training process.
The embedding is interactively connected to the respective images, and basic selection and filtering options, as in the literature~\cite{prasad2024tree}, are provided.
The training process can be paused upon detecting anomalies or biases that could compromise the model's integrity. 
For example, training can be stopped if latent vectors or generated data proceed in an undesirable trajectory.
This pause allows corrections, including training datasets or the model architecture, to mitigate detected biases. 
In this paper, we focus on data augmentation for fair representations. 
For example, if some groups are poorly performing or underrepresented, data augmentation can be one way to fix it. 
In our method, we propose automatically scraping the web using Google Custom Search APIs for required images with appropriate queries.
Post-intervention, training may be resumed or restarted, depending on the extent of the issues. 
Analyses of model representations and data distributions across subsequent iterations after corrections are tangible proof of the model's adjustments. 
\newline
\newline
Progressively analyzing training evolution with our framework facilitates the early detection of biases, optimizes resource utilization by enabling timely interventions, and fosters transparent model development.

\section{Analysing the training process of GANs}
To demonstrate our frameworks' effectiveness, we applied it to AttentionGAN~\cite{tang2021attentiongan}, a well-known model for unpaired image-to-image translation. 
We trained the model to change human hair color to black, blond, brown, or grey with the CelebA dataset~\cite{celeba}. 
Human data is particularly relevant since the impartiality and accuracy of the generated data are crucial.
AttentionGAN operates by taking an input face image and a target domain (i.e., hair color), aiming to change the color accurately. The discriminator differentiates between real and generated images and classifies the hair color. The generator aims to fool the discriminator while also optimizing for accurate domain classification. Additionally, the generator incorporates a cyclic L1 reconstruction loss for consistency.

We employed the original hyperparameters~\cite {tang2021attentiongan} but modified the target domains to hair colors. The training process, by default, involved 200,000 iterations. 
2,000 samples were randomly selected for validation, and their hair color was transformed for analysis.
In addition to the standard loss metrics, we evaluated the model's performance and potential biases by extracting generated images and discriminator feature vectors every 5000 iterations. 
Discriminator features here refer to the max-pooled activations of the layer before the logits layer.
These model elements were used to analyze the model's efficacy and fairness.

\subsection{Biases with young grey-haired women}
\begin{figure}[t!]
    \centering
    \includegraphics[width=0.99\linewidth]{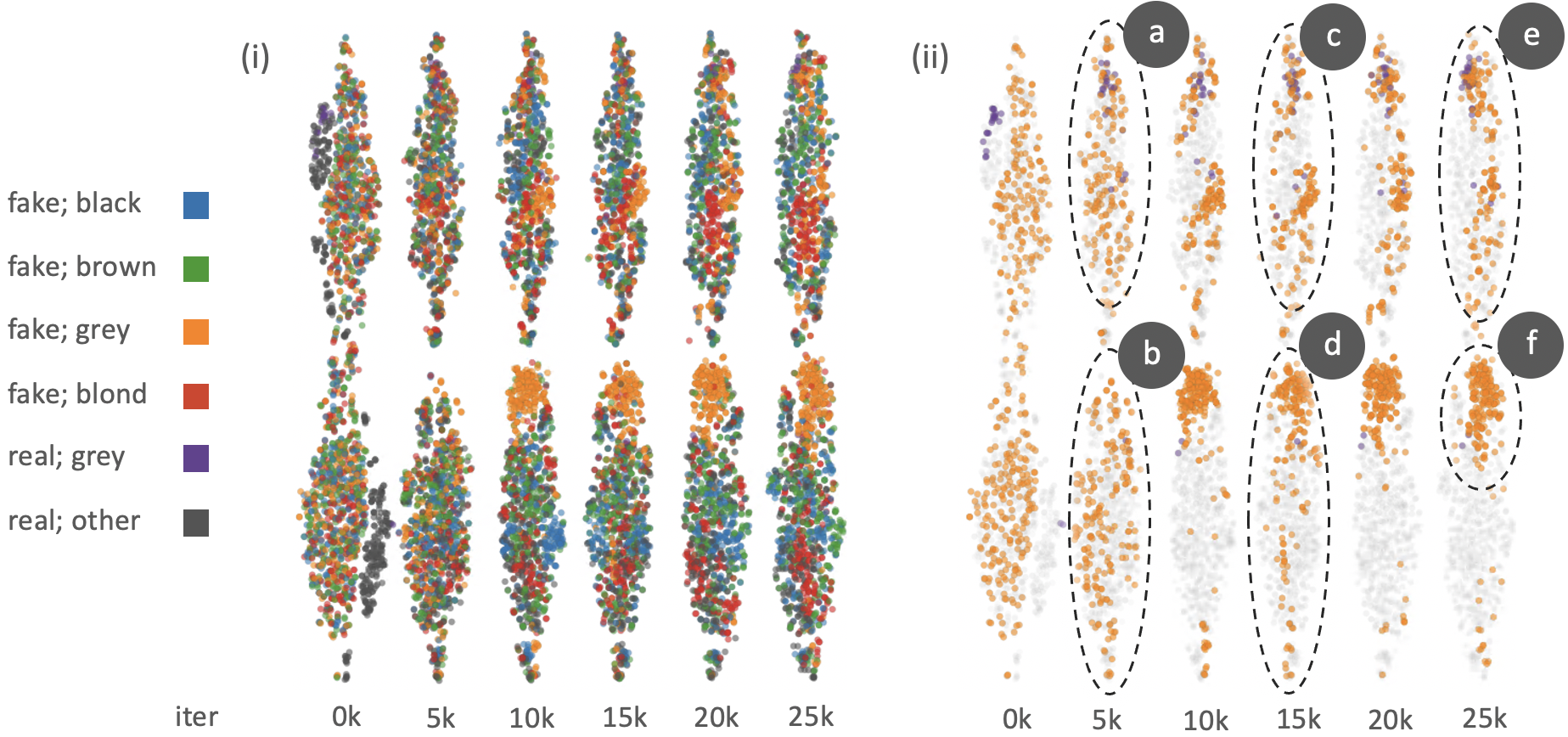}
    \caption{(i) The left panel shows the evolution of CLIP image embeddings across images. 
    Interaction between the embeddings and images reveals that the orange points correspond to generated grey instances. 
    These grey points are initially mixed with other hair color data until iteration 5000 (clusters a and b). They become more distinct after (clusters c through f).
    Filtering only grey instances (ii) shows two clusters: one that includes a mix of real grey instances (cluster e) and another that is distinct (cluster f).}
    \label{fig:nogrey_evol_10k}
\end{figure}
\begin{figure}[b!]
    \centering
    \includegraphics[width=0.99\linewidth]{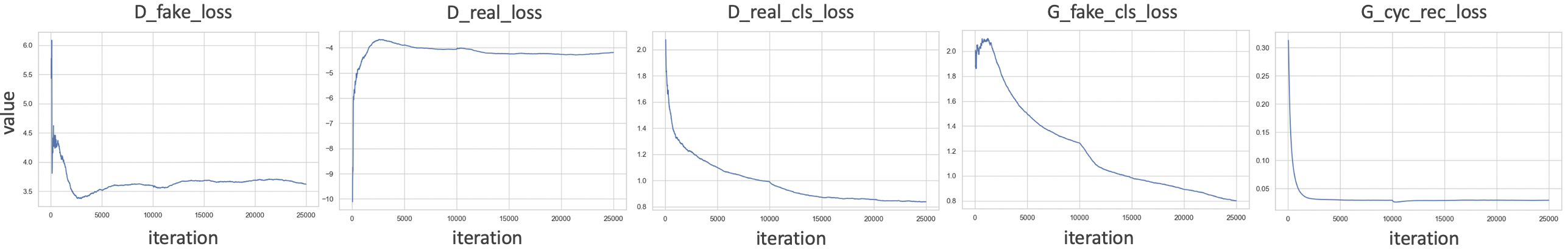}
    \caption{Aggregated loss metrics, including (from left to right) real vs. fake image losses, classification cross-entropy losses for real and fake images, and the cyclic L1 reconstruction loss across 25000 training iterations.}
    \label{fig:lossmetrics}
\end{figure}
Figure~\ref{fig:nogrey_evol_10k} represents the evolution of the CLIP (image encoder) embeddings of the real and fake images over the first 25,000 iterations.
Initially (first iteration in Figure~\ref{fig:nogrey_evol_10k}), the embeddings of real and generated images are dispersed, indicating that the generator is still producing unrealistic images. 
Progressively, these real and fake points are mixed, indicative of the generator refining its output to more closely mimic real images, as indicated by converging loss metrics such as reconstruction L1 loss, domain classification cross-entropy, and the discriminator's ability to distinguish real from fake images (see Figure~\ref{fig:lossmetrics}).
Despite these improvements in loss metrics, significant biases become apparent in our embedding analysis, especially in depicting young grey-haired women.

Two distinct groups based on gender are observed in the embeddings: the top one predominantly represents men (see Figure~\ref{fig:nogrey_evol_10k}a), while the bottom represents women (see Figure~\ref{fig:nogrey_evol_10k}b), as differentiated by CLIP. Within these gender clusters, hair colors appear mixed, indicating a limitation of CLIP in distinguishing hair color due to its small proportion of the overall image.

\begin{figure}[b!]
    \centering
    \includegraphics[width=0.99\linewidth]{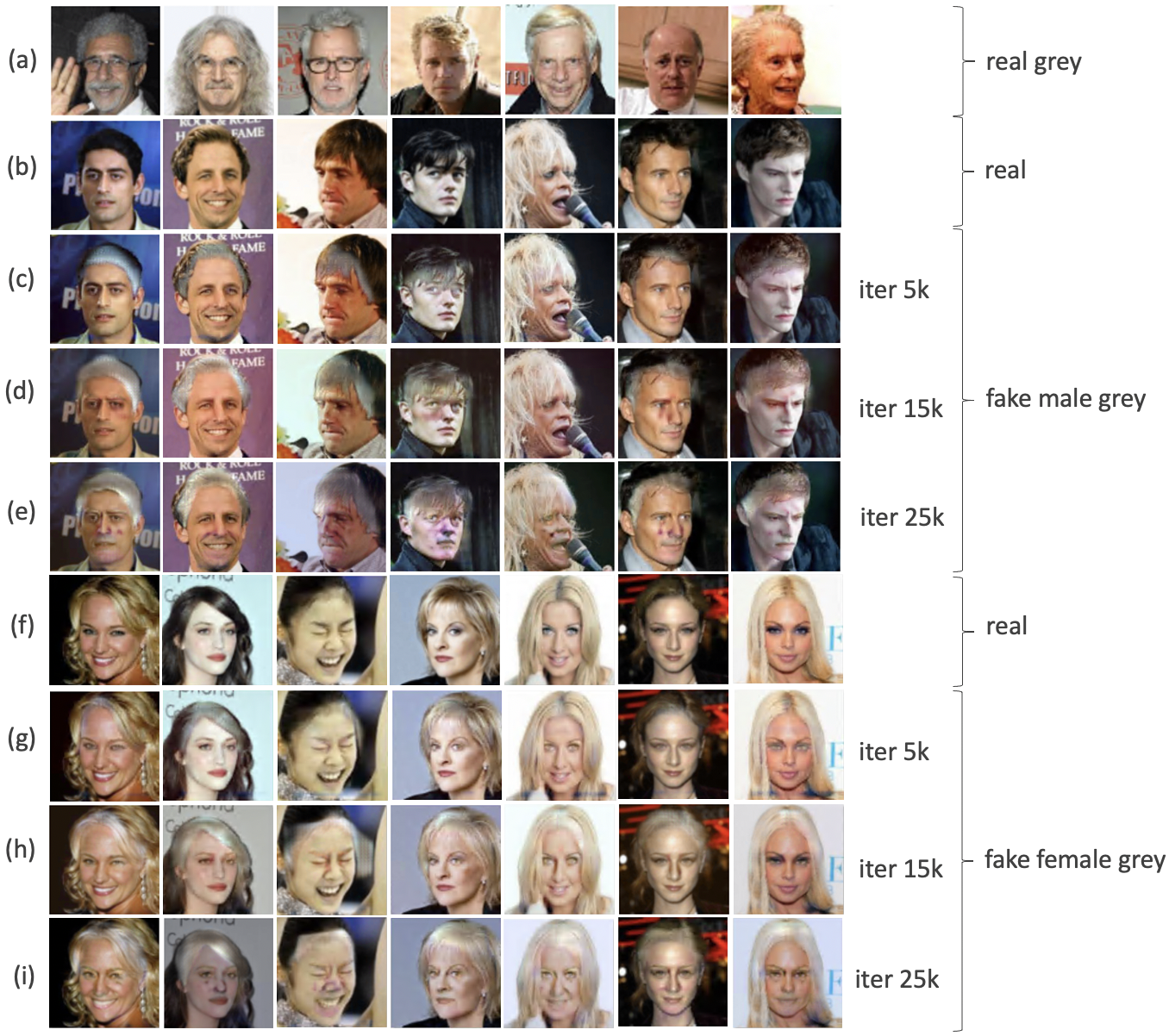}
    \caption{The real (a) and generated grey-haired images for men (c-e) and women (g-i). The evolution from iteration 5000 (c and g) to 15000 (d and h) and 25000 (e and i) reveals exacerbated effects on facial features. Initially, at iteration 5000, the hair color starts to change, but from iteration 15000, the aging effects become more pronounced. Women (g-i) exhibit more significant facial distortions.}
    \label{fig:nogrey_imgevol_10k}
\end{figure}
After 5,000 iterations, a distinct clustering of grey hair represented by orange instances emerges, separating from other hair colors (see Figure~\ref{fig:nogrey_evol_10k}d).
This clustering becomes more pronounced as training progresses (see Figure~\ref{fig:nogrey_evol_10k}f). 
This separate clustering, specifically for grey hair, potentially indicates significant changes in the images that differ from the other colors.
When examining the distribution of real grey-haired points (in purple), it becomes evident that the generated grey-haired faces, particularly those of women, do not overlap with the real ones.
This lack of overlap indicates that the generator struggles to produce realistic grey-haired images for women.
Inspecting actual images illustrates that real grey-haired samples mainly consist of older men in Figure~\ref{fig:nogrey_imgevol_10k}a. This bias is reflected in the generated images, where changing hair color to grey modifies the perceived age. 
While the model initially focuses on changing the hair color (see Figures~\ref{fig:nogrey_imgevol_10k}c, g), as the training progresses, iterations show more significant distortions in the facial features (see Figures~\ref{fig:nogrey_imgevol_10k}d, h) that increase over iterations (see Figures~\ref{fig:nogrey_imgevol_10k}e, i). 
The generated faces of women are particularly distorted, aligning with the embedding analysis.
The hair color of women initially changes correctly (see Figure~\ref{fig:nogrey_imgevol_10k}g), reflected in the grey hair cluster overlapping with other hair colors in iteration 5000 of Figure~\ref{fig:nogrey_evol_10k}b. 
However, in later iterations, there are additional alterations of facial features (see Figure~\ref{fig:nogrey_evol_10k}f and Figure~\ref{fig:nogrey_imgevol_10k}i). 
While some of these effects are also seen in male faces in Figure~\ref{fig:nogrey_imgevol_10k}e, they are less pronounced, as also reflected by less apparent orange clusters in Figure~\ref{fig:nogrey_evol_10k}e.
\begin{figure}[h!]
    \centering
    \includegraphics[width=0.95\linewidth]{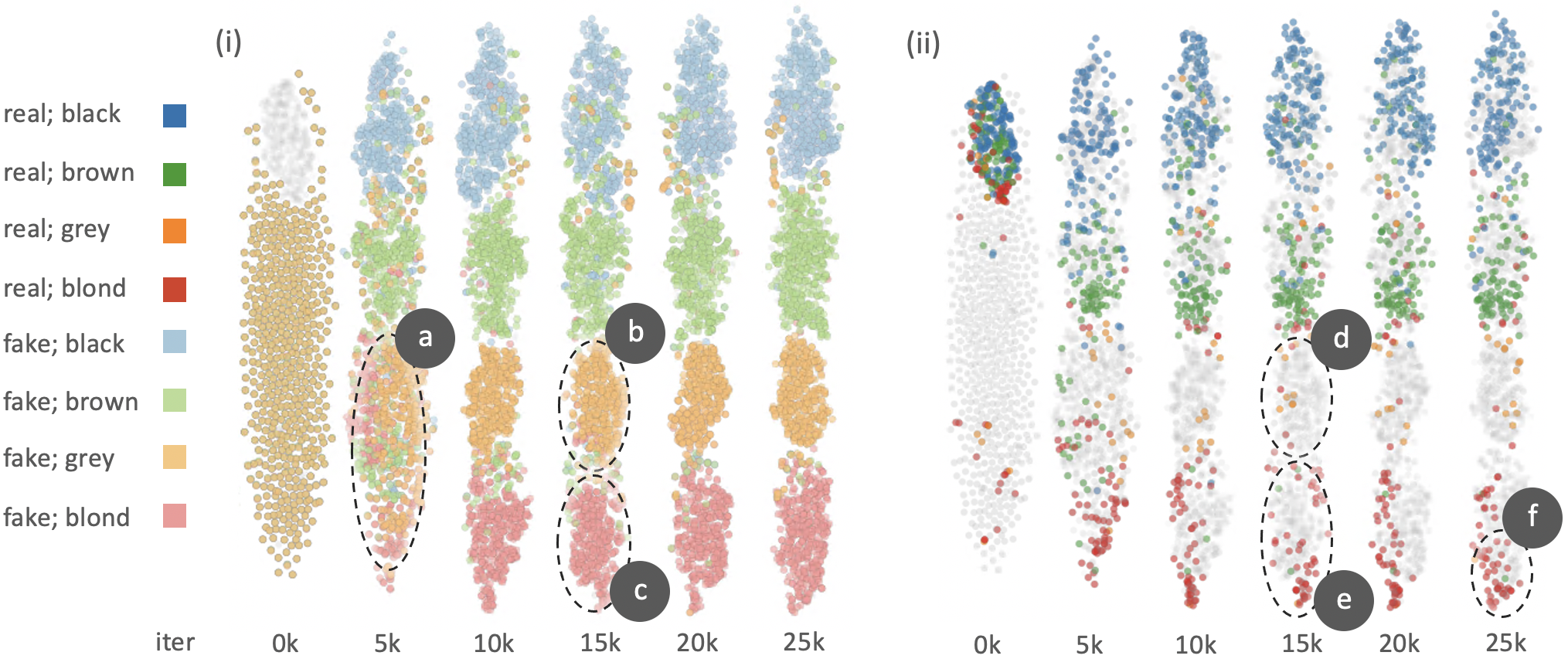}
    \caption{The evolution of the discriminator's final feature vector for all images. Initially, feature vectors of fake grey images (light orange) and fake blond images (light pink) are mixed until iteration 5000 (a). They gradually separate in iterations (b) and (c). Real images are represented by fewer points with minimal overlap, especially for grey-haired (d) and partially blond-haired (e, f) samples.}
    
    \label{fig:nogrey_evol_10k_feats}
\end{figure}

Discriminator analysis is crucial for understanding how well it distinguishes between real and generated images and classifies hair colors, including grey. By examining the feature space of the discriminator, we can assess whether the model is accurately producing realistic images or relying on misleading attributes.
Figure~\ref{fig:nogrey_evol_10k_feats} provides insight into this discriminator's feature space evolution. 
Initially, feature vectors for grey fake and blond fake images are intermixed (see Figure~\ref{fig:nogrey_evol_10k_feats}a), gradually separating as training progresses (see Figure~\ref{fig:nogrey_evol_10k_feats}b, c). This separation suggests improved discrimination but correlates with observed changes in facial features of generated grey-haired images (see Figure~\ref{fig:nogrey_imgevol_10k}d, h), indicating a potential bias in the model's optimization direction toward older faces rather than changing the hair color.
Similar to the CLIP embedding analysis in Figure~\ref{fig:nogrey_evol_10k_feats}, real grey-haired images (depicted in dark orange) exhibit minimal overlap with their corresponding fake samples, underscoring the persistence of the issue.
Both the discriminator and generator aim to minimize hair color classification loss. Given the distribution issues in the training data for grey hair, it is logical that the model relies on the most significant discriminative attributes (in this case, age) to minimize losses.

\subsection{Biases with blond-haired men}
\begin{figure}[t!]
    \centering
    \includegraphics[width=0.9\linewidth]{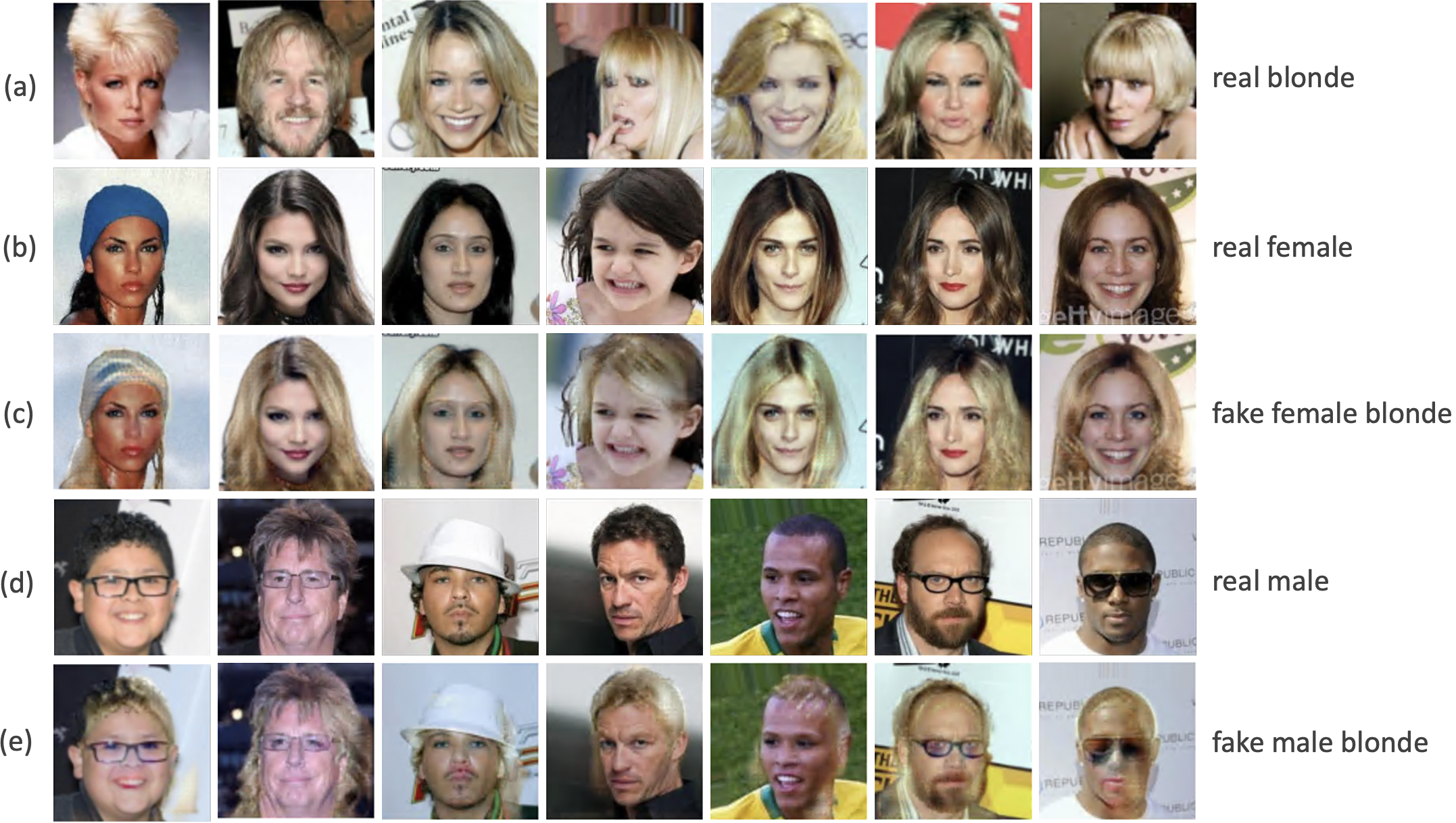}
    \caption{The figure shows images generated at iteration 25000. (a) Displays original images of real blond-haired individuals. Real female images (b) are depicted with desired transformations to blond hair (c). In contrast, (d-e) illustrates real male faces undergoing a feminization process, including longer hair and a change in hair color.}
    
    \label{fig:nogrey_imgevol_blond}
\end{figure}
While blond and grey hair clusters become more distinct in the feature space (see Figure~\ref{fig:nogrey_evol_10k_feats}b, c), the analysis of real image distributions reveals minimal overlap for grey-haired images and partial overlap for blond-haired images.
Interestingly, the partial overlap primarily in Figure~\ref{fig:nogrey_evol_10k_feats}f involves blond female faces, indicating good quality generated images.
Examining fake male blond faces reveals deviations towards more feminine features, diverging from expected male characteristics (see Figure~\ref{fig:nogrey_imgevol_blond}e). 
Real blond images, predominantly featuring female faces (see Figure~\ref{fig:nogrey_imgevol_blond}a), highlight biases in the model's training data, resulting in alterations such as increased hair length and enhanced lip brightness in male faces to optimize variations between real and fake blonde images.
These biases towards older men with grey hair and women with blond hair underscore challenges in the generator's ability to capture essential facial features accurately across genders and hair colors. These insights extend beyond performance metrics, revealing critical limitations in the model's generative capabilities and inherent biases.

This comprehensive analysis gives us a deeper understanding of the model's strengths and weaknesses in realistic image synthesis across varied demographics and hair color distributions early in the generation process for timely corrections. 

\subsection{Fixing biases} 
\begin{figure}
    \centering
    \includegraphics[width=0.8\linewidth]{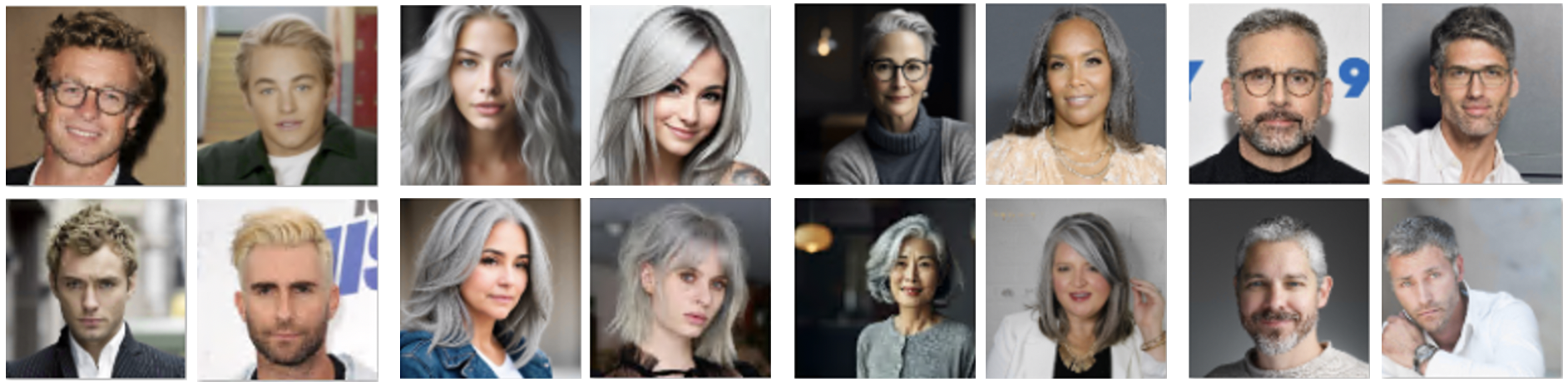}
    \caption{Images scraped from Google for (left to right) blond males, young grey-haired women, grey-haired women, and grey-haired men.}
    \label{fig:augimgs}
\end{figure}
Upon identifying significant biases in the training data concerning grey and blond hair, we paused the training process to investigate and rectify these issues. 
Our analysis revealed a lack of diverse samples, particularly in categories such as young people with grey hair and blond-haired males. 
Hence, we utilized Google APIs to automatically source more images, specifically focusing on "young woman with grey hair," "young man with grey hair," "woman with grey hair," "man with grey hair," and "man with blond hair" (see Figure~\ref{fig:augimgs}). 
After a high-level manual cleanup of these images, we used a stratified sampling strategy to create a more balanced dataset, aiming to mitigate the observed biases.
This led to approximately 650 blond males, 500 grey females, and 500 grey male images, on which standard augmentation methods such as rotations and translations were applied to extend them further.

Following these data augmentations, we resumed training from iteration 5000, where initial biases were noted. By enhancing the diversity and representativeness of our training data, we expect the model to produce better outputs, leading to improved performance and reduced biases in image synthesis.

\subsection{Performance and quality improvements}
\begin{figure}[t!]
    \centering
    \includegraphics[width=0.99\linewidth]{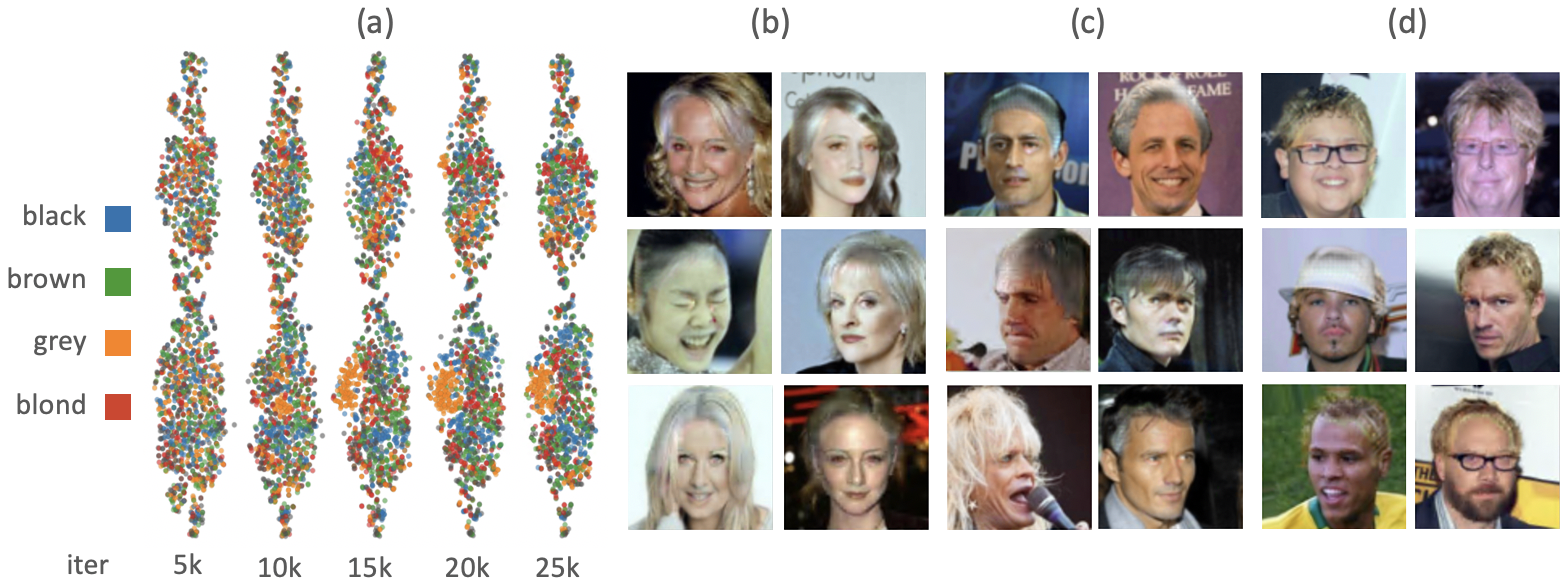}
    \caption{Improved data generation with the model trained with augmented datasets. (a) The evolution of CLIP features, continuing from iteration 5000 onward, shows that the greys are mixed more with the other data points. The new images of grey-haired women (b), men (c), and blond-haired men (d) at iteration 25000. }
    \label{fig:newgreyblond_evol_10k}
\end{figure}

\begin{figure}[t!]
    \centering
    \includegraphics[width=0.9\linewidth]{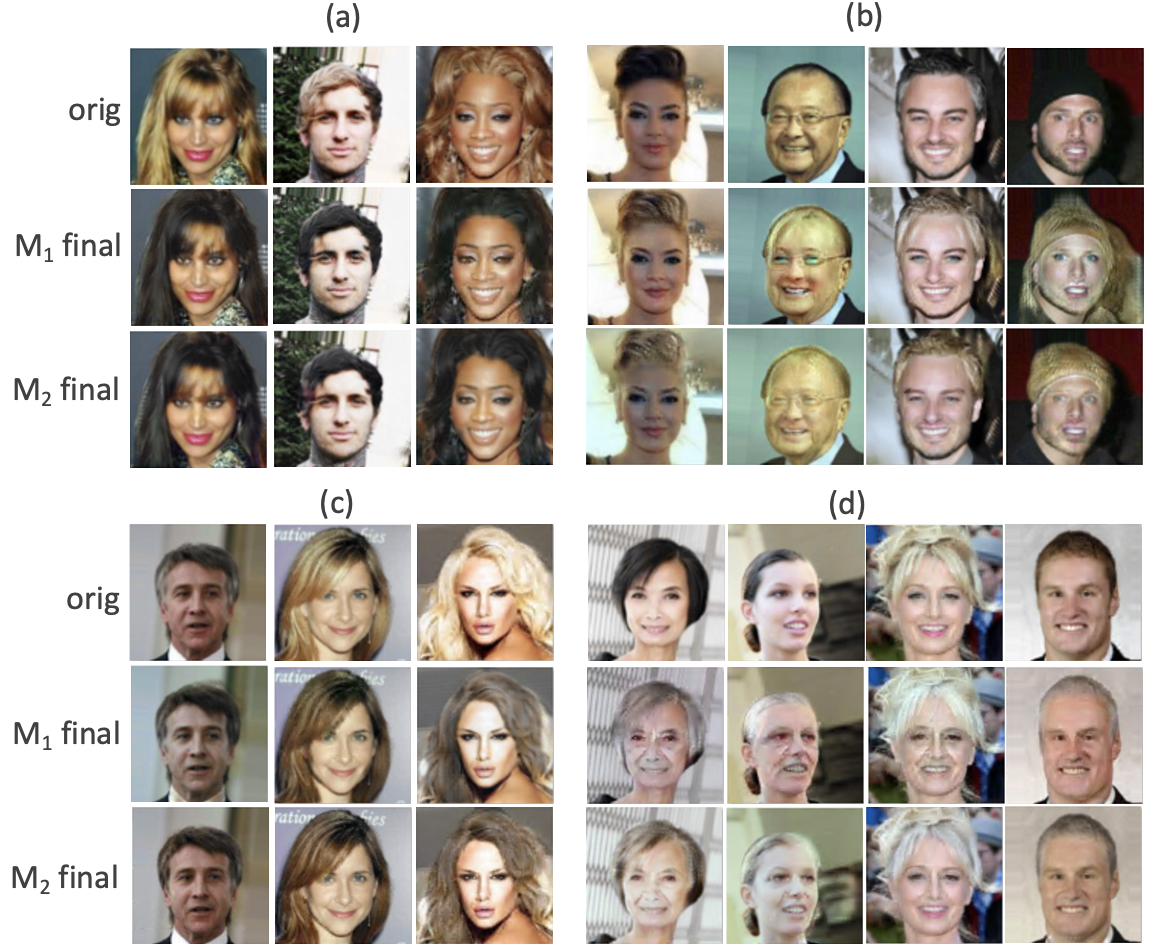}
    \caption{Final generated data of the new model with augmentations ($M_2$) vs. the original without augmentations if the original training had been completed ($M_2$) and original samples across changing hair to black (a), blond (b), brown (c), and grey (d).  }
    
    \label{fig:improve_final}
\end{figure}

After implementing data augmentation and resuming training from iteration 5000, we observed notable enhancements in our model's performance and quality. Consistent with our previous methodology, we evaluated the CLIP image embedding space using the same set of 2000 validation samples at every $n^{th}$ iteration.
The updated embeddings revealed a significant change in the clustering of samples. 
In the earlier model, as depicted in Figure~\ref{fig:nogrey_evol_10k}f, a distinct cluster of grey samples, representing generated female grey-haired images, was separated from other hair color clusters. 
However, this grey hair cluster is no longer prominent in the new model. 
The samples are now more interspersed with other colors, indicating a reduction in bias (see Figure~\ref{fig:newgreyblond_evol_10k}).
\newline
\newline
\noindent
\textbf{Evolution variations:} Visual inspection of generated images at iteration 25,000 revealed that grey-haired individuals no longer appeared aged, and blond-haired males did not exhibit modified features such as longer hair in Figure~\ref{fig:newgreyblond_evol_10k}b, c, d, addressing the earlier biases in the original generated images in Figure~\ref{fig:nogrey_imgevol_10k}. 
While there are still some changes, for example, the pink lip colors for the darker-skinned blond males in Figure~\ref{fig:newgreyblond_evol_10k}d, it is still an improved result over the initially generated images.
Further availability of specific metadata or nuanced image encoders could help this analysis.
\newline
\newline
\noindent
\textbf{Qualitative analysis:} On exploring the what-if final outputs of the original model without data augmentation ($M_1$), they displayed biases for grey females and blond males (see Figure~\ref{fig:improve_final}d and Figure~\ref{fig:improve_final}b), showing significant aging and feminine effects, respectively. 
On the other hand, the augmented model's outputs were more realistic across genders and hair colors (see Figures~\ref{fig:improve_final}d and~\ref{fig:improve_final}b). 
Performance of black and brown transformations remained consistent in Figures~\ref{fig:improve_final}a and Figures~\ref{fig:improve_final}c.
\newline
\newline
\noindent
\textbf{Quantitative analysis:}
We check the quality of the generated images with the widely used Frechet Inception Distance (FID), which measures the similarity between the final real and generated datasets.
FID was checked per hair color within the 2000-sample validation set.
For $M_1$, we used the final images that would have been generated if the full training had been completed.
FID for blond and grey hair colors improved from $77.20$, $40.99$ for $M_1$ to $82.4$, $47.30$ for $M_2$.
\newline
\newline
\noindent
\textbf{Computational costs:} 
Finally, this training intervention and correction improved model performance and saved considerable training time by addressing biases early rather than through new retraining.  Specifically, by pausing the original training process at 25,000 out of 200,000 iterations, we only used 12.5\% of the computational resources for the first run, significantly reducing the time and resources needed compared to identifying biases post-training and starting over.
\newline
\newline
While the improvements for addressing biases in facial attributes beyond hair colors were primarily driven by data augmentation, another effective strategy could be incorporating additional domains in the discriminator's classifier. 
This would ensure that other features do not change unintentionally, although it would require additional labels for various facial attributes.
Overall, our approach demonstrates the importance of early bias detection and correction, leading to a more efficient and effective training process and resulting in higher-quality outputs from the generative model.

\section{Conclusion}
In this work, we tackle the critical issues of bias and inefficiency in training deep generative models (DGMs) by proposing a progressive analysis framework. 
Given the growing complexity and computational demands of DGMs, there is a growing need for real-time monitoring and mid-training interventions. 

Our framework addresses this need by providing an interpretable method for analyzing pattern changes or the evolution of latent space encodings and generated distributions across training iterations.
By focusing on the progression of high-dimensional model data, we show how our framework supports identifying and mitigating biases related to gender and age in hair color transformations using a GAN. Our results demonstrate that timely interventions improve model fairness and accuracy while conserving computational resources. This approach provides a means to enhance the training process of generative models, leading to more reliable and unbiased outcomes.

Looking ahead, the concept of real-time embedding analysis for bias and error evolution presented here can be extended beyond GANs to other generative models, such as diffusion models in the future. 

%
%
\bibliographystyle{splncs04}
\bibliography{main}

\begin{thebibliography}{10}
\providecommand{\url}[1]{\texttt{#1}}
\providecommand{\urlprefix}{URL }
\providecommand{\doi}[1]{https://doi.org/#1}

\bibitem{bau2018visualizing}
Bau, D., Zhu, J.Y., Strobelt, H., Zhou, B., Tenenbaum, J.B., Freeman, W.T., Torralba, A.: Visualizing and understanding generative adversarial networks. In: International Conference on Learning Representations (2019)

\bibitem{bond2021deep}
Bond-Taylor, S., Leach, A., Long, Y., Willcocks, C.G.: Deep generative modelling: A comparative review of vaes, gans, normalizing flows, energy-based and autoregressive models. IEEE transactions on pattern analysis and machine intelligence  \textbf{44}(11),  7327--7347 (2021)

\bibitem{jeong2022interactively}
Jeong, S., Liu, S., Berger, M.: Interactively assessing disentanglement in gans. In: Computer Graphics Forum. vol.~41, pp. 85--95. Wiley Online Library (2022)

\bibitem{jin2020generative}
Jin, L., Tan, F., Jiang, S.: Generative adversarial network technologies and applications in computer vision. Computational intelligence and neuroscience  \textbf{2020}(1),  1459107 (2020)

\bibitem{tcav}
Kim, B., Wattenberg, M., Gilmer, J., Cai, C., Wexler, J., Viegas, F., et~al.: Interpretability beyond feature attribution: Quantitative testing with concept activation vectors (tcav). In: International conference on machine learning. pp. 2668--2677. PMLR (2018)

\bibitem{liu2017analyzing}
Liu, M., Shi, J., Cao, K., Zhu, J., Liu, S.: Analyzing the training processes of deep generative models. IEEE transactions on visualization and computer graphics  \textbf{24}(1),  77--87 (2017)

\bibitem{celeba}
Liu, Z., Luo, P., Wang, X., Tang, X.: Deep learning face attributes in the wild. In: Proceedings of International Conference on Computer Vision (ICCV) (December 2015)

\bibitem{luccioni2024stable}
Luccioni, S., Akiki, C., Mitchell, M., Jernite, Y.: Stable bias: Evaluating societal representations in diffusion models. Advances in Neural Information Processing Systems  \textbf{36} (2024)

\bibitem{van2008visualizing}
Van~der Maaten, L., Hinton, G.: Visualizing data using t-sne. Journal of machine learning research  \textbf{9}(11) (2008)

\bibitem{manduchi2024challenges}
Manduchi, L., Pandey, K., Bamler, R., Cotterell, R., D{\"a}ubener, S., Fellenz, S., Fischer, A., G{\"a}rtner, T., Kirchler, M., Kloft, M., et~al.: On the challenges and opportunities in generative ai. arXiv preprint arXiv:2403.00025  (2024)

\bibitem{mcinnes2018umap}
McInnes, L., Healy, J., Melville, J.: Umap: Uniform manifold approximation and projection for dimension reduction. arXiv preprint arXiv:1802.03426  (2018)

\bibitem{morales2024efficient}
Morales-Juarez, E., Fuentes-Pineda, G.: Efficient generative adversarial networks using linear additive-attention transformers. arXiv preprint arXiv:2401.09596  (2024)

\bibitem{conceptevo}
Park, H., Lee, S., Hoover, B., Wright, A.P., Shaikh, O., Duggal, R., Das, N., Li, K., Hoffman, J., Chau, D.H.: Concept evolution in deep learning training: A unified interpretation framework and discoveries. In: Proceedings of the 32nd ACM International Conference on Information and Knowledge Management. CIKM '23 (2023)

\bibitem{park2023content}
Park, J., Son, S., Lee, K.M.: Content-aware local gan for photo-realistic super-resolution. In: Proceedings of the IEEE/CVF International Conference on Computer Vision. pp. 10585--10594 (2023)

\bibitem{pezzotti2017deepeyes}
Pezzotti, N., H{\"o}llt, T., Van~Gemert, J., Lelieveldt, B.P., Eisemann, E., Vilanova, A.: Deepeyes: Progressive visual analytics for designing deep neural networks. IEEE transactions on visualization and computer graphics  \textbf{24}(1),  98--108 (2017)

\bibitem{prasad2024tree}
Prasad, V., van Gorp, H., Humer, C., Vilanova, A., Pezzotti, N.: The tree of diffusion life: Evolutionary embeddings to understand the generation process of diffusion models. arXiv preprint arXiv:2406.17462  (2024)

\bibitem{prasad2022transform}
Prasad, V., van Sloun, R.J., van~den Elzen, S., Vilanova, A., Pezzotti, N.: The transform-and-perform framework: Explainable deep learning beyond classification. IEEE Transactions on Visualization and Computer Graphics  \textbf{30}(2),  1502--1515 (2022)

\bibitem{clip}
Radford, A., Kim, J.W., Hallacy, C., Ramesh, A., Goh, G., Agarwal, S., Sastry, G., Askell, A., Mishkin, P., Clark, J., et~al.: Learning transferable visual models from natural language supervision. In: International conference on machine learning. pp. 8748--8763. PMLR (2021)

\bibitem{tang2021attentiongan}
Tang, H., Liu, H., Xu, D., Torr, P.H., Sebe, N.: Attentiongan: Unpaired image-to-image translation using attention-guided generative adversarial networks. IEEE transactions on neural networks and learning systems  \textbf{34}(4),  1972--1987 (2021)

\bibitem{wang2018ganviz}
Wang, J., Gou, L., Yang, H., Shen, H.W.: Ganviz: A visual analytics approach to understand the adversarial game. IEEE transactions on visualization and computer graphics  \textbf{24}(6),  1905--1917 (2018)

\bibitem{wu2022sustainable}
Wu, C.J., Raghavendra, R., Gupta, U., Acun, B., Ardalani, N., Maeng, K., Chang, G., Aga, F., Huang, J., Bai, C., et~al.: Sustainable ai: Environmental implications, challenges and opportunities. Proceedings of Machine Learning and Systems  \textbf{4},  795--813 (2022)

\bibitem{yang2023disdiff}
Yang, T., Wang, Y., Lv, Y., Zheng, N.: Disdiff: Unsupervised disentanglement of diffusion probabilistic models. arXiv preprint arXiv:2301.13721  (2023)

\bibitem{yosinski2015understanding}
Yosinski, J., Clune, J., Nguyen, A., Fuchs, T., Lipson, H.: Understanding neural networks through deep visualization. arXiv preprint arXiv:1506.06579  (2015)

\bibitem{yu2020inclusive}
Yu, N., Li, K., Zhou, P., Malik, J., Davis, L., Fritz, M.: Inclusive gan: Improving data and minority coverage in generative models. In: Computer Vision--ECCV 2020: 16th European Conference, Glasgow, UK, August 23--28, 2020, Proceedings, Part XXII 16. pp. 377--393. Springer (2020)

\bibitem{zhao2021codegan}
Zhao, J., Liu, Z., Guo, X., Pan, L.: Codegan: Contrastive disentanglement for generative adversarial network. arXiv e-prints pp. arXiv--2103 (2021)

\bibitem{zhou2021evaluation}
Zhou, S.: On the Evaluation of Deep Generative Models. Stanford University (2021)

\end{thebibliography}
\end{document}